\documentclass[runningheads]{llncs}
\usepackage[T1]{fontenc}
\IfFileExists{newtxtext.sty}{\usepackage{newtxtext}\usepackage[varvw]{newtxmath}}{\usepackage{txfonts}}
\usepackage{graphicx}
\usepackage{array}
\usepackage[hidelinks]{hyperref}
\providecommand{\Description}[1]{}
\newcommand{\us}{u_{\mathrm{s}}}
\newcommand{\ue}{u_{\mathrm{e}}}
\newcommand{\tc}{\tau_c}
\newcommand{\expl}{$^{\dagger}$}
\newcolumntype{L}[1]{>{\raggedright\arraybackslash}p{#1}}
\newcolumntype{R}[1]{>{\raggedleft\arraybackslash}p{#1}}

\begin{document}
\title{Uncertainty Evaluation to Support Safety of the Intended Functionality Analysis for Determining the Applicability of Performance Targets Across Triggering Conditions in ML-Based LiDAR Object Detection}
\titlerunning{Applicability of Performance Targets Across Triggering Conditions}
\author{Milin Patel\inst{1}\orcidID{0000-0002-8357-6018} \and
	Rolf Jung\inst{2}\orcidID{0000-0002-0366-4844}}
\authorrunning{M. Patel and R. Jung}
\institute{Institute for Driver Assistance and Connected Mobility,\\
	Kempten University of Applied Sciences, Benningen, Germany
	\and
	Faculty of Computer Science,\\
	Kempten University of Applied Sciences, Kempten, Germany\\
	\email{\{milin.patel, rolf.jung\}@hs-kempten.de}}
\maketitle

\begin{abstract}
For safety of the intended functionality (SOTIF) analysis according to ISO~21448, a machine learning object detector for light detection and ranging (LiDAR) point clouds can meet a performance target, a limit on the mean per-frame share of wrong detections among accepted detections, by rejecting detections above an uncertainty threshold. A threshold selected on data of one condition carries a statistical guarantee only for data distributed like its selection data, whereas SOTIF analysis covers every triggering condition. This paper presents an uncertainty evaluation in which the threshold is transferred into a triggering condition, a test decides at a stated error probability whether the target is applicable there, not applicable or not determined, and a lower bound counts the independent units an applicable outcome requires. For PointPillars ensembles on KITTI and nuScenes, counted in recorded sequences, as the dependence between frames requires, neither dataset held enough independent units for any target limit, and no transfer had a not applicable outcome. Counted in frames, the same data gave not applicable outcomes at limits fixed in advance after transfers into far distance, into rain and between two separately trained detectors, thresholds that met the target limit after a transfer rejected most correct detections, and in rain the area under the receiver operating characteristic curve did not decrease. A performance target met under one condition therefore counts as evidence under a triggering condition only after a test there, reported with its share of accepted correct detections and its number of independent recorded sequences.
\end{abstract}

\section{Introduction}\label{sec:intro}

An automated vehicle perceives its surroundings with object detectors, among them detectors for point clouds from light detection and ranging (LiDAR) sensors. A detector based on machine learning (ML) outputs a three-dimensional box, called a detection, for each object that it recognises in a frame, one point cloud of the sensor. The motion planner, which plans the path and speed of the vehicle, acts on these detections. This paper considers detections of cars, the object class with the most annotations in the KITTI and nuScenes datasets used here~\cite{geiger2012kitti,caesar2020nuscenes}. A detection is wrong if it does not match a car marked by the annotators of the dataset (an annotated car), for example because no object lies in its box or because the box misses the position or size of the car. A wrong detection can make the vehicle brake or steer for an object that does not exist, or misjudge the position of one that does.

ISO~21448, the standard on the safety of the intended functionality (SOTIF), requires that the residual risk of a driving function meets acceptance criteria at vehicle level~\cite{iso21448}. Performance targets for system elements, for example a limit on how often a sensor reports an object that does not exist, can be derived from these criteria. A detector can meet such a target by rejecting uncertain detections. In this paper, the detector is an ensemble of separately trained networks, called \emph{members}, and two indicators compute the uncertainty of a detection from the outputs of the members. The \emph{spatial indicator} $\us$ grows as the members disagree more on the position and size of the box, and the \emph{existence indicator} $\ue$ grows as their confidence that an object is present falls. A detection whose indicator value exceeds the \emph{uncertainty threshold} $\tau$ is rejected and does not reach the motion planner. The other detections are accepted. The performance target of this paper bounds the \emph{false acceptance proportion} (FAP), the share of wrong detections among the accepted detections of a frame averaged over frames, by a limit $\alpha$. A lower threshold rejects more wrong detections but also more correct ones, and every rejected correct detection is a car that the motion planner does not receive.

Learn then Test (LTT) selects a threshold on recorded data so that the FAP does not exceed $\alpha$ with probability at least $1-\delta$, where $\delta$ is a chosen error probability~\cite{angelopoulos2021ltt}. LTT makes no assumption about the form of the data distribution, but it assumes that the recorded units, for example frames, are independent draws from one distribution. The guarantee therefore holds only for data distributed like the data on which the threshold was selected~\cite{tibshirani2019,barber2023}, with their mix of distances, weather and locations.

ISO~21448 describes a \emph{triggering condition} (TC) as a specific condition of a scenario that initiates a system reaction contributing to a hazardous behaviour, and it requires the response of the system to TCs to be evaluated. Its example of a statistical argument (Clause~C.6.3.2) assumes one environment and does not state what holds in another. ISO/PAS~8800 considers the case in which the safety requirements of an artificial intelligence system are fulfilled only for a limited part of the input space. It requires that such a case is identified and reported to the development of the encompassing system~\cite{iso8800}. A SOTIF analysis therefore has to establish under which conditions a performance target is met. Whether a threshold selected under one condition meets the target under another has to be tested.

The conference paper extended here~\cite{patel2026vehits} reported the area under the receiver operating characteristic (ROC) curve (AUROC). AUROC measures how well an indicator separates correct from wrong detections (0.5 is chance level, 1 is perfect separation), not whether a threshold meets a limit. It is therefore not known whether AUROC shows a change of condition under which the threshold no longer meets the limit.

Recorded data can show that a limit is met only if they hold enough independent units (frames or recorded sequences). Frames of one \emph{recorded sequence}, a continuous recording such as a nuScenes scene or a KITTI drive, are not independent of each other. The number of independent frames or recorded sequences needed to show that a limit is met sets the minimum amount of data to record per condition.

\begin{figure}[!t]
\centering
\includegraphics[width=\textwidth]{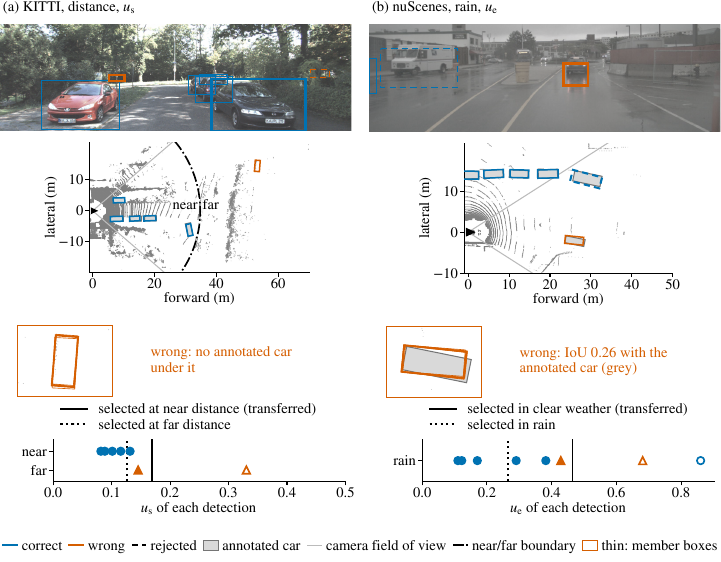}
\Description{Two columns. (a) A KITTI frame: camera image with the detection boxes, a bird's-eye view of the LiDAR points with the annotated cars, the detections and the near/far boundary, an inset with the member boxes of the accepted wrong detection, and a strip with the spatial indicator of every detection and both thresholds. (b) A nuScenes frame in rain with the same layout, the existence indicator and the thresholds selected in clear weather and in rain. In both columns the transferred threshold accepts one wrong detection that the threshold selected in the condition of use rejects.}
\caption{A wrong detection that the transferred threshold accepts and the threshold selected in the condition of use rejects. (a)~KITTI, far distance, spatial indicator $\us$, threshold transferred from near distance. (b)~nuScenes, rain, existence indicator $\ue$, threshold transferred from clear weather. Bottom strips: indicator value of each detection. Open markers: rejected by the transferred threshold. IoU: intersection over union. Both thresholds were selected at the exploratory limit 0.1\expl, which defines no performance target.}
\label{fig:teaser}
\end{figure}

In the uncertainty evaluation of this paper, a statistical test, the \emph{applicability test}, decides whether a performance target met under one condition is also met under another condition. A threshold is \emph{transferred} when it is selected on recorded data of one condition, the \emph{condition of selection}, and used unchanged on recorded data of another, the \emph{condition of use}. For example, a threshold selected on the detections at near distance is transferred when it is used unchanged on the detections at far distance. At a stated error probability, the performance target is \emph{applicable} in the condition of use if the data show that the FAP with the transferred threshold does not exceed the limit there, and \emph{not applicable} if they show that it exceeds the limit. The applicability is \emph{not determined} if the data show neither. The test examines whether the target is met under the condition of use, not whether it is required there. This paper takes the target as required under every condition of the operational design domain (ODD). Figure~\ref{fig:teaser} shows, for two transfers, one recorded frame in which the transferred threshold accepts a wrong detection that the threshold selected in the condition of use rejects. Both frames were chosen to show this case. In the KITTI frame, an unannotated object under the box of the wrong detection cannot be excluded.

\subsection{Research Questions}\label{sec:rq}
This paper examines three changes of condition in both directions: between near and far distance, between clear weather and rain, and between the KITTI and nuScenes datasets. A \emph{condition} is near or far distance, clear weather or rain, or one dataset together with its trained detector, assigned to the car detections by the rules of Sect.~\ref{sec:classes}. Far distance and rain are treated as potential TCs, and near distance and clear weather are not. Two datasets differ in the sensor, the location, the mix of objects and, here, the trained detector at once, so a change of dataset does not isolate a TC. It serves as an experimental control, against which the changes of distance and of weather can be compared. This paper answers three research questions (RQs).
\begin{description}
\item[RQ1] Does an uncertainty threshold that keeps the false acceptance proportion, the share of wrong detections among the accepted detections averaged over frames, within a required limit under the condition for which it was selected keep it within this limit under another condition, such as the potential triggering conditions far distance and rain?
\item[RQ2] When an uncertainty threshold no longer keeps the false acceptance proportion within the required limit under another condition, which changes in the detections from the condition for which it was selected accompany this, and which of these changes does the area under the receiver operating characteristic curve (AUROC) show?
\item[RQ3] How many independent recorded frames or sequences (continuous recordings, such as nuScenes scenes and KITTI drives) are needed to show that an uncertainty threshold keeps the false acceptance proportion within a required limit under a given condition, such as a potential triggering condition, and which limits can the KITTI and nuScenes datasets show to be met?
\end{description}

\subsection{Contributions and Scope}\label{sec:contrib}
\begin{enumerate}
\item The \emph{applicability test}, which decides at a stated error probability whether a performance target met with an uncertainty threshold under one condition is applicable, not applicable or not determined under another, together with the measure that follows from each outcome in the SOTIF analysis and a lower bound on the number of independent units needed to show that the target is met (Sects.~\ref{sec:method} and~\ref{sec:implications}).
\item An evaluation of the test with ensembles of PointPillars networks on KITTI and nuScenes after changes of object distance and weather and, as a control, of dataset and detector, with the frame and with the recorded sequence as the independent unit, including the share of correct detections that each threshold with an applicable outcome accepts (RQ1, Sect.~\ref{sec:res-rq1}).
\item An identification of the changes in the detections that accompany each not applicable outcome, which shows that AUROC does not reveal the changes found in rain and between the datasets (RQ2, Sect.~\ref{sec:res-rq2}).
\item A determination of the limits that the frames and the recorded sequences of KITTI and nuScenes can show to be met (RQ3, Sect.~\ref{sec:res-rq3}).
\end{enumerate}
The test combines known statistical elements: the LTT selection, a confidence interval computed on data not used for the selection, and the zero-failure bound on the number of units. The contribution lies in their use for the SOTIF question of a transfer across conditions and in the measured outcomes for a LiDAR detector. The scope is the object detector as one system element. This paper proposes no new uncertainty indicator, detector or threshold selection method. It derives no value of $\alpha$ from an acceptance criterion and does not estimate how often each condition occurs in operation, because both belong to the argument at vehicle level.

\paragraph{Relation to the conference paper.} The conference paper~\cite{patel2026vehits} evaluated uncertainty indicators of an ensemble LiDAR detector for the analysis activities of ISO~21448 and applied a rule that rejects uncertain detections. It used point clouds from a driving simulator and ensemble outputs simulated by perturbation, not by trained networks, and it selected the thresholds of the rule on all frames, without separate data to test them. This paper uses ensembles trained on two recorded datasets and keeps apart the data that fix the candidate thresholds, select a threshold and test it. It selects thresholds with a finite-sample guarantee and tests them in the condition of use. The test, the measurements and the results do not appear in the conference paper.

\section{Related Work}\label{sec:related}

\subsection{Performance Targets in SOTIF Processes}\label{sec:rw-targets}
ISO~21448 defines the \emph{acceptance criterion} as a criterion representing the absence of an unreasonable level of risk and the \emph{validation target} as a value used to argue that the criterion is met~\cite{iso21448}, and it uses \emph{performance targets} for system elements without defining the term. Hofmeister and Kinalzyk trace acceptance criteria to performance targets in an ISO~21448 process~\cite{hofmeister2023}, Madala et al.\ list the factors behind acceptance criteria and validation targets~\cite{madala2022}, and Reisgys et al.\ decompose quantitative risk acceptance criteria into verification and validation targets~\cite{reisgys2025}. Kramer et al.\ identify and quantify hazardous scenarios~\cite{kramer2020}, and Hoss et al.\ review the testing of object-based perception~\cite{hoss2022}. Argument patterns for SOTIF and for ML components~\cite{birch2020,hawkins2021amlas} and the treatment of uncertainty in ML assurance~\cite{burton2023uncertainty} define where evidence for a system element enters a safety argument. None of these works examines whether a target met under one condition is also met under another.

\subsection{Statistical Risk Control and Its Validity Under Distribution Shift}\label{sec:rw-risk}
Selective classification lets a model abstain on some inputs and bounds the risk of the predictions that it makes~\cite{geifman2017selective}. Conformal risk control~\cite{angelopoulos2022crc} bounds the expected value of a chosen error rate. Risk-controlling prediction sets~\cite{bates2021rcps} and LTT~\cite{angelopoulos2021ltt} keep this expected value below a limit with a stated probability for the number of samples at hand. All three methods select a parameter on calibration data held out for that purpose. Conformal selection controls the false discovery rate over a selected set~\cite{jin2022selection}. These guarantees hold on average over the distribution of the calibration data. Selecting the parameter separately inside each subset of the data gives a guarantee per subset~\cite{vovk2005,vovk2012cond}, which becomes uninformative when a subset holds few data points~\cite{ding2023classcond}. Under covariate shift, reweighting the data by the ratio of the two input densities restores the guarantee~\cite{tibshirani2019}, and when the data are not exchangeable, the loss of the guarantee can still be bounded~\cite{barber2023}. Conformal prediction and conformal risk control have been applied to camera-based object detection with calibration and test data from one dataset~\cite{andeol2023,timans2024}. This paper uses LTT without extending it and tests, for a LiDAR detector, whether a threshold selected with LTT meets its limit after a transfer to another condition.

\subsection{Uncertainty of LiDAR Object Detectors}\label{sec:rw-lidar}
The indicators used here are computed from the members of a deep ensemble, several separately trained networks~\cite{lakshminarayanan2017}. The regression uncertainty of object detectors has been evaluated with proper scoring rules~\cite{harakeh2021}. The calibration of predictive uncertainty degrades under dataset shift~\cite{ovadia2019}, and three-dimensional detectors trained in one country lose accuracy in another~\cite{wang2020traingermany}. None of these works tests a threshold across conditions.

\subsection{Condition-Specific Evidence for Perception}\label{sec:rw-condition}
Salay et al.\ decompose a misperception rate over disjoint perception conditions by the law of total probability~\cite{salay2022missinglink}, and CUBICS weights Bayesian performance estimates per operating situation with the expected frequency of each situation~\cite{herd2026cubics}. Kumar et al.\ combine a conformal prediction set with an abstention threshold selected once on one calibration set and report the abstention rate for each corruption severity~\cite{kumar2025dualthreshold}. Lee et al.\ adapt the confidence threshold of a LiDAR detector to object distance~\cite{lee2024adaptive}. Li et al.\ select the confidence threshold for each synthetic sensor degradation so that the share of wrong detections among the accepted ones stays below a limit~\cite{li2026bgc}. Adee et al.\ model LiDAR performance limitations as a Bayesian network to search for TCs~\cite{adee2021edcc,adee2021icsrs}. None of these works, and no other study in LiDAR object detection found in a documented search (supplementary material), tests whether a threshold selected under one condition meets its limit under another (Table~\ref{tab:nearest}).

\begin{table}[t]
\centering
\caption{Works closest to the question whether a threshold selected under one condition meets its limit under another.}
\label{tab:nearest}
\footnotesize
\begin{tabular}{|L{19mm}|L{40mm}|L{13mm}|L{19mm}|L{17mm}|}
\hline
work & quantity that is set or estimated & set or estimated per condition & limit tested in another condition & finite-sample guarantee\\
\hline
Salay~\cite{salay2022missinglink} & misperception rate, estimated & yes & no & no\\
\hline
CUBICS~\cite{herd2026cubics} & performance per situation, Bayesian estimate weighted by situation frequency & yes & no & no (Bayesian)\\
\hline
Kumar~\cite{kumar2025dualthreshold} & abstention threshold, selected once & no (reported per severity) & no & no (coverage of the prediction set only)\\
\hline
Lee~\cite{lee2024adaptive} & confidence threshold, adapted to distance & yes & no & no\\
\hline
Li~\cite{li2026bgc} & confidence threshold, per sensor degradation & yes & no & no\\
\hline
this paper & uncertainty threshold, selected with LTT in each condition & yes & yes, tested in the condition of use & yes (LTT, independent units)\\
\hline
\end{tabular}
\end{table}

\section{Test of the Applicability of a Performance Target Across Conditions}\label{sec:method}

\begin{figure}[t]
\centering
\includegraphics[width=\textwidth]{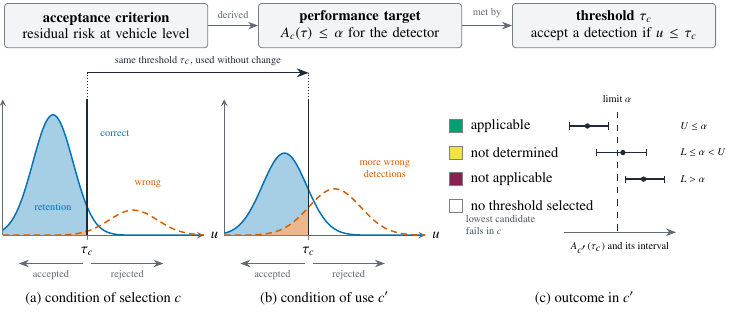}
\Description{Top: three boxes linked by arrows, from the acceptance criterion at vehicle level to the performance target A_c(tau) <= alpha for the detector to the threshold tau_c. Below, panel (a) shows schematic distributions of the uncertainty of correct detections (blue) and wrong detections (orange, dashed) in condition c with a vertical threshold line. An arrow carries the same threshold to panel (b), condition c', where wrong detections are more frequent and a larger orange area lies left of the line. Panel (c) shows four outcome rows: an interval of the false acceptance proportion left of the dashed limit (applicable), across it (not determined), right of it (not applicable), and a row without interval (no threshold selected).}
\caption{The applicability test. (a)~Threshold $\tc$ selected in the condition of selection $c$, for example near distance. (b)~The same threshold in the condition of use $c'$, for example far distance. (c)~Outcome in $c'$ from the interval $[L,U]$ of the FAP, or no threshold selected if the lowest candidate threshold fails in $c$. $u$: indicator value; $A_c(\tau)$: FAP in $c$. The distributions are schematic.}
\label{fig:concept}
\end{figure}

The applicability test takes a threshold selected with a guarantee in the condition of selection $c$ and uses it unchanged in the condition of use $c'$. From an interval of the FAP in $c'$, it decides whether the performance target is applicable, not applicable or not determined (Fig.~\ref{fig:concept}). It takes as given a trained detector with its uncertainty indicators, the limit $\alpha$ derived from the acceptance criterion at vehicle level and the error probability $\delta$ chosen in the SOTIF analysis. Table~\ref{tab:symbols} lists the main symbols.

\begin{table}[t]
\centering
\caption{Symbols.}
\label{tab:symbols}
\footnotesize
\setlength{\tabcolsep}{3pt}
\begin{tabular}{|L{11mm}|L{44mm}|L{12mm}|L{45mm}|}
\hline
$c$, $c'$ & condition of selection, condition of use & $\tau$ & uncertainty threshold\\
$f$, $f_c$ & frame; car detections of $c$ in $f$ & $\tau_j$ & candidate threshold\\
$d$, $y(d)$ & detection; 1 if correct, 0 if wrong & $\tc$ & threshold selected in $c$\\
$M$, $m(d)$ & number of members of the ensemble; number of members that contribute a box to $d$ & $a_f$ & per-frame share of false acceptances among the accepted detections in $f_c$\\
$b_k$, $s_k$ & box of member $k$; its confidence & $A_c$, $\hat{A}_c$ & FAP in $c$; its estimate\\
$\us$, $\ue$ & spatial and existence indicator & $A$, $\pi_c$ & FAP over all frames with a car detection; share of them in $c$\\
$u(d)$ & indicator value of $d$ & $N_f$ & number of accepted detections in $f_c$\\
$\alpha$ & limit & $n$ & number of independent units\\
$\delta$ & error probability & $L$, $U$ & lower and upper confidence bound of $A_{c'}(\tc)$\\
$\delta'$ & $\delta$ for selecting, $\delta/2$ for testing a threshold & & \\
\hline
\end{tabular}
\end{table}

\subsection{Uncertainty Indicators and the False Acceptance Proportion}\label{sec:quantity}
The detector is an ensemble of $M$ members, networks of one architecture trained from different random seeds. A detection $d$ merges the boxes $b_k$ of the $m(d)$ members that contribute a box to it ($3\le m(d)\le M$, Sect.~\ref{sec:detectors}), and $s_k$ is the confidence of member $k$ for its box. The two indicators are taken from the conference paper~\cite{patel2026vehits}, and Eq.~(\ref{eq:us}) is adapted from it. With the intersection over union (IoU) of two rotated three-dimensional boxes, which is 1 for identical and 0 for disjoint boxes, the spatial indicator is
\begin{equation}\label{eq:us}
\us(d)=1-\frac{2}{m(d)\,(m(d)-1)}\sum_{k<l}\mathrm{IoU}(b_k,b_l),
\end{equation}
where $k$ and $l$ run over the contributing members. The existence indicator is $\ue(d)=1-\frac{1}{M}\sum_{k=1}^{M}s_k$, with $s_k=0$ for a member that contributes no box.

A detection is \emph{correct} ($y(d)=1$) if one-to-one matching assigns it to an annotated car with an IoU of at least 0.5, and \emph{wrong} ($y(d)=0$) otherwise. A wrong detection can be a box where no object is, a box on an annotated car that misses the overlap criterion, a second box on a car already matched, or a box on an object of another class or without annotation. A detection is \emph{accepted} if its indicator value $u(d)$, either $\us(d)$ or $\ue(d)$, does not exceed the threshold $\tau$, and \emph{rejected} otherwise. An accepted wrong detection is called a \emph{false acceptance}, and a rejected correct detection is called \emph{withheld}. For a condition $c$, the share of false acceptances among the accepted detections of $c$ in frame $f$ (the \emph{per-frame share}) and its expectation over the frames of $c$ (the FAP) are
\begin{equation}\label{eq:fap}
a_f(\tau)=\frac{\bigl|\{d\in f_c:\ u(d)\le\tau,\ y(d)=0\}\bigr|}{\bigl|\{d\in f_c:\ u(d)\le\tau\}\bigr|},
\qquad A_c(\tau)=\mathbb{E}\bigl[a_f(\tau)\mid f\in c\bigr],
\end{equation}
where $f_c$ denotes the car detections of condition $c$ in frame $f$, a frame $f$ belongs to $c$ ($f\in c$) if $f_c$ is not empty, and a frame without accepted detections in $f_c$ contributes $a_f=0$. The \emph{performance target} is $A_c(\tau)\le\alpha$, and $\alpha$ is called the \emph{limit}. The estimate of $A_c$ on a set of frames is written $\hat{A}_c$. In terms of object detection metrics, $a_f$ is one minus the precision of the accepted detections of a frame, and $A_c$ is the false discovery rate computed per frame and averaged over frames~\cite{angelopoulos2021ltt}. The FAP differs from the \emph{pooled share}, the share of wrong detections among all accepted detections of a condition taken together. The per-frame form is used because LTT bounds a mean of per-unit values in $[0,1]$. It gives every frame the same weight.

Because a frame of $c$ without accepted detections contributes $a_f=0$,
\begin{equation}\label{eq:trivial}
A_c(\tau)=P\bigl(N_f(\tau)>0\bigr)\;\mathbb{E}\bigl[a_f(\tau)\mid N_f(\tau)>0\bigr]\;\le\;P\bigl(N_f(\tau)>0\bigr),
\end{equation}
where $N_f(\tau)$ is the number of accepted detections in $f_c$, and $P$ refers to the frames of $c$. If $P(N_f(\tau)>0)<\alpha$, the limit is met even if every accepted detection is wrong. Every applicable outcome is therefore reported with the \emph{retention}, the share of the correct detections of a condition that the threshold accepts, estimated on its test part (Sect.~\ref{sec:parts}).

\subsection{Selection of a Threshold per Condition}\label{sec:threshold}
The frames of each condition are divided into a selection part, used to select thresholds, and a test part, used to test them (Sect.~\ref{sec:parts}). A candidate threshold $\tau_j$, fixed in advance, passes if the Hoeffding--Bentkus (HB) $p$-value of the selection data of $c$ for the null hypothesis $A_c(\tau_j)>\alpha$ is at most $\delta$. For $n$ selection frames of $c$ with mean $\hat{A}_c(\tau_j)$ of $a_f(\tau_j)$, this $p$-value is \[p=\min\bigl\{\exp\bigl(-n\,h(\min(\hat{A}_c(\tau_j),\alpha),\alpha)\bigr),\ e\,P\bigl(\mathrm{Bin}(n,\alpha)\le\lceil n\hat{A}_c(\tau_j)\rceil\bigr)\bigr\}\] with a binomial random variable $\mathrm{Bin}(n,\alpha)$, Euler's number $e$ and $h(v,w)=v\ln(v/w)+(1-v)\ln((1-v)/(1-w))$, where $0\ln0=0$~\cite{angelopoulos2021ltt}. This $p$-value is valid for any distribution of the per-frame shares $a_f$ in $[0,1]$. The count $n$ includes frames without accepted detections, and the $p$-value requires independent and identically distributed units.

The candidates are taken in order from the lowest upwards, starting where $A_c$ is small by Eq.~(\ref{eq:trivial}). The selected threshold $\tc$ is the last candidate that passes before the first candidate that fails. If the lowest candidate fails, no threshold is selected in $c$. The FAP $A_c$ need not increase monotonically with $\tau$, because a lower threshold can remove a correct detection and keep a wrong one. A candidate above the first failure can therefore pass, but it is not selected. If the selection frames of $c$ are independent draws from the frame distribution of $c$, this fixed order bounds by $\delta$ the probability that any candidate passing before the first failure has an FAP above $\alpha$~\cite{angelopoulos2021ltt}. The limit therefore holds for $\tc$ with probability at least $1-\delta$, for the distribution of $c$ only.

If every frame with a car detection belongs to exactly one condition and $\pi_c$ is the share of these frames in condition $c$, then
\begin{equation}\label{eq:mix}
A(\tau)=\sum_c\pi_c A_c(\tau)\le\max_c A_c(\tau).
\end{equation}
A bound on the FAP $A$ over all frames with a car detection therefore refers to the mix $\pi$ of the recording campaign, which need not equal the mix in operation. Weather and dataset assign each frame to exactly one condition. For them, a threshold that meets $\alpha$ in every condition meets it under every mix, provided that the frames within each condition are distributed in operation as in the recorded data. Distance is assigned per detection, and a frame can hold near and far detections, so this premise fails for distance.

\subsection{Decision on Applicability in the Condition of Use}\label{sec:rule}
A threshold $\tc$ is tested on the test part of condition $c'$. The upper confidence bound $U$ of $A_{c'}(\tc)$ is obtained by inverting the HB $p$-value at error probability $\delta/2$, and the lower confidence bound $L$ is one minus the same bound computed from the values $1-a_f$, so $[L,U]$ is an interval at confidence level $1-\delta$. The performance target is \emph{applicable} in $c'$ if $U\le\alpha$, \emph{not applicable} if $L>\alpha$, and \emph{not determined} if $L\le\alpha<U$. If no threshold can be selected in $c$, the outcome is \emph{no threshold selected} (Fig.~\ref{fig:concept}c). For a given true FAP, only one of the two erroneous outcomes is possible, so the test gives an erroneous applicable or not applicable outcome with probability at most $\delta/2$. This holds if the test units are independent draws from $c'$ and independent of the selection units. This error probability applies to each outcome on its own, so a safety argument that relies on several outcomes at once has to divide $\delta$ by their number. A not determined outcome makes no claim, but it can hide an exceeded limit.

Used in its own condition of selection ($c'=c$) with independent frames, a threshold gives the outcome not applicable only after an erroneous selection (probability at most $\delta$) or an erroneous interval (at most $\delta/2$). The probability of this outcome is therefore at most $\delta+\delta/2$, and its observed frequency serves as a check of the selection (Sect.~\ref{sec:res-rq1}).

\subsection{Changes in the Detections That Can Lead to an Exceeded Limit}\label{sec:causes}
A transferred threshold is the same value on the scale of the indicator in both conditions. The pooled share that it gives in a condition depends on the share of wrong detections among all car detections of the condition and on the distributions of the indicator for correct and for wrong detections. The FAP, a mean over frames, depends in addition on how detections are spread over frames. Four changes in the detections from $c$ to $c'$ can therefore lead to an exceeded limit.
\begin{description}
\item[Weaker separation.] The two distributions overlap more in $c'$.
\item[More wrong detections.] A larger share of wrong detections in $c'$ at unchanged distributions.
\item[Shifted values.] The values of both kinds of detection lie at another position on the scale, so the same threshold accepts a different share of them.
\item[Different spread over frames.] The wrong detections are distributed differently over the frames, for example concentrated in frames with few accepted detections.
\end{description}
AUROC measures separation only. It does not depend on the share of wrong detections~\cite{davis2006,saito2015} or on the spread of detections over frames, and it is unaffected by any rescaling of the indicator that keeps the order of the values. The ROC curve plots, for every threshold, the share of the wrong detections rejected against the share of the correct detections withheld, and a threshold marks its \emph{operating point} on this curve. A threshold that marks the same operating point in $c$ and $c'$ rejects the same shares of the correct and of the wrong detections in both conditions. An exceeded limit then comes from more wrong detections or a different spread over frames. Weaker separation and more wrong detections lower the retention that any threshold on this indicator can reach at the limit in $c'$, whereas shifted values are compensated by a threshold selected in $c'$.

\subsection{Minimum Number of Independent Units to Show That a Limit Is Met}\label{sec:units}
A \emph{unit} is a frame or a recorded sequence that the test treats as one independent draw. Units without any false acceptance are the data most favourable to a passing candidate and to the outcome applicable. If every per-unit value is 0 or 1 and the true FAP lies just above $\alpha$, all of $n$ independent units are free of false acceptances with a probability close to $(1-\alpha)^n$. Showing that the limit is met at error probability $\delta'$ therefore requires at least
\begin{equation}\label{eq:samplesize}
n\ \ge\ \frac{\ln\delta'}{\ln(1-\alpha)}
\end{equation}
independent units per condition and part, with $\delta'=\delta$ for selecting and $\delta'=\delta/2$ for testing a threshold. This lower bound is the zero-failure bound, about $3/\alpha$ at $\delta'=0.05$ (the rule of three). It holds for every test that is valid for any distribution of the per-unit values in $[0,1]$ and in which fewer false acceptances never weaken the evidence. The HB test is such a test. The bound is necessary, not sufficient, and any unit with a false acceptance raises the number further. It does not apply to the outcome not applicable, which a few units with many false acceptances can show.

\section{Detectors, Datasets and Conditions of the Study}\label{sec:setup}

The test is applied to ensembles of PointPillars networks on KITTI and nuScenes for the three changes of condition of Sect.~\ref{sec:rq}.

\subsection{Ensemble Detectors and Rule for Merging Member Boxes}\label{sec:detectors}
Each ensemble consists of $M=4$ PointPillars networks~\cite{lang2019pointpillars}, trained from different random seeds with the OpenPCDet toolbox~\cite{openpcdet2020} (version 0.6.0). On KITTI~\cite{geiger2012kitti}, the members use the three-class configuration of the toolbox and are trained for 10 passes (epochs) over the 3{,}712 training frames. On nuScenes~\cite{caesar2020nuscenes}, the members use the class-balanced multi-head configuration~\cite{zhu2019cbgs} and are trained for about 10 of its 20 epochs on the 700 training scenes. Each member keeps the car boxes with a confidence of at least 0.1. The boxes of the first member are taken in descending order of confidence, and each other member contributes its unused car box with the highest IoU with that box, if the IoU is at least 0.5. A detection is formed only if at least three of the four members contribute a box, a rule called the \emph{vote}. The merged box is the mean of the member boxes, and its confidence is the mean confidence of the contributing members.

The vote is itself a rejection rule, applied before any uncertainty threshold. Table~\ref{tab:setup} lists the share of all annotated cars, without a distance or visibility filter, that a detection after the vote matches (the recall after the vote). Retention therefore refers to the correct detections after the vote, not to all annotated cars. Because a missing member contributes $s_k=0$, a detection with three members has $\ue\ge0.25$, and a threshold on $\ue$ below 0.25 accepts only detections found by all four members.

\subsection{Assignment of Distance, Weather and Dataset Conditions}\label{sec:classes}
Far distance and rain are treated as potential TCs because both act on the point cloud: rain affects the returns through droplets in the air and water on the road, and an object at greater distance returns fewer points. This paper does not show that they initiate a hazardous behaviour. Weather and dataset labels are assigned per frame by rules that use no output of the detector.

Every detection is assigned the distance of its box centre from the LiDAR origin in the ground plane, because a wrong detection has no annotated car whose distance could be used. Distance therefore describes where the detector places a box: the far condition is defined by the detector output, not by the distance of an annotated car. The near/far boundary is the median distance of the car detections in the grid part (Sect.~\ref{sec:parts}). A frame contributes its near detections to the near condition and its far detections to the far condition, so near and far share frames.

A nuScenes scene~\cite{caesar2020nuscenes} is a rain scene if its text description contains the word `rain' and a night scene if it contains the word `night'. Night scenes are excluded, so that clear and rain frames do not also differ in lighting. The daylight scenes without rain are called clear scenes. The validation set holds 24 rain scenes in daylight. The label gives no rain intensity and does not separate falling rain from a wet road. Clear scenes can include a road wet from earlier rain, and clear and rain scenes also differ in location and traffic.

KITTI marks image regions that are not annotated (DontCare) and treats vans as a neighbouring class of cars in its benchmark~\cite{kitti_devkit}. A detection in a DontCare region or on a van is therefore not necessarily wrong. KITTI results are reported with these 2{,}454 detections excluded (\emph{KITTI rule}), and Sect.~\ref{sec:res-rq1} reports the effect of including them.

The change of dataset and detector uses two ensembles, the \emph{KITTI detector} trained on KITTI and the \emph{nuScenes detector} trained on nuScenes. In the transfer from nuScenes to KITTI, a threshold selected with the nuScenes detector on nuScenes is used with the KITTI detector on KITTI, and the reverse transfer works in the same way. The test therefore examines the transfer of a threshold value between two separately trained detectors, not for one detector into another environment. Distance is examined on both datasets, so the three changes of condition give the four pairs of conditions of Table~\ref{tab:setup}.

\begin{table}[t]
\centering
\caption{Pairs of conditions. Last column: frames per condition in the selection and test parts of split~0 (one of five random assignments of the frames to the parts), first line for the first condition. Boundaries: split~0.}
\label{tab:setup}
\footnotesize
\begin{tabular}{|L{19mm}|L{38mm}|L{25mm}|R{20mm}|}
\hline
pair of conditions & recorded frames and car detections & conditions & frames, selection / test\\
\hline
KITTI distance & 3{,}769 validation frames; 27{,}175 car detections (KITTI rule); 0.858 of 14{,}385 annotated cars matched & near / far, boundary 34.8\,m & 1{,}484 / 1{,}500 \newline 1{,}514 / 1{,}545\\
\hline
nuScenes distance & 6{,}019 validation frames from 150 scenes; 40{,}036 car detections; 0.398 of 65{,}465 annotated cars matched & near / far, boundary 21.4\,m & 2{,}207 / 2{,}209 \newline 2{,}270 / 2{,}276\\
\hline
nuScenes weather & 4{,}449 clear and 968 rain frames, all in daylight; 602 night frames excluded & clear / rain & 1{,}917 / 1{,}900 \newline 392 / 414\\
\hline
change of dataset and detector & all validation frames of KITTI and of nuScenes & KITTI with the KITTI detector (KITTI rule) / nuScenes with the nuScenes detector & 1{,}844 / 1{,}852 \newline 2{,}849 / 2{,}850\\
\hline
\end{tabular}
\end{table}

\subsection{Division of the Frames and Choice of the Independent Unit}\label{sec:parts}
\begin{figure}[t]
\centering
\includegraphics[width=\textwidth]{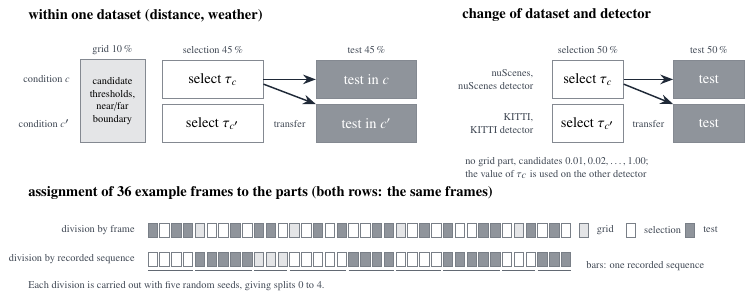}
\Description{Left: within one dataset, a grid part fixes the candidate thresholds and the near/far boundary; a threshold selected on the selection part of condition c goes by arrow to the test part of c and of c'. Right: for the change of dataset and detector, each dataset has its own ensemble and is divided into halves; the threshold selected on nuScenes goes to the test part of KITTI. Bottom: the same 36 frames assigned to grid, selection and test part, once by frame and once by recorded sequence.}
\caption{Division of the recorded frames into parts within one dataset (left) and for the change of dataset and detector (right). Arrows: a threshold selected in $c$ is tested in $c'$ (one transfer direction) and in $c$ (check of the selection). Bottom: 36 example frames divided by frame or by sequence.}
\label{fig:parts}
\end{figure}

Within one dataset, a grid part (10\,\%) fixes the near/far boundary and the grid of candidate thresholds, the quantiles of the indicator at levels $0.01,0.02,\dots,1.00$ over the car detections of both conditions. A selection part (45\,\%) selects the thresholds, and a test part (45\,\%) tests them (Fig.~\ref{fig:parts}). For the change of dataset and detector, the frames of each dataset are divided into halves, and the candidates $0.01, 0.02, \dots, 1.00$ are fixed in advance, so that neither dataset influences them. Every division is carried out with five random seeds, and each resulting assignment of the frames to the parts is called a \emph{split}. The splits divide the same recorded frames in different ways. An outcome on all five splits therefore has the same bound on the error probability as an outcome on one split and shows only that it does not depend on the split.

Frames of one recorded sequence, a nuScenes scene of about 20\,s or a KITTI drive, share objects, weather and road. Consecutive annotated nuScenes frames are 0.5\,s apart and show the same objects from nearly the same position. ISO/PAS~8800 lists such nearly identical frames, one in the training and the next in the test data, as a lack of independence (Table~11-2)~\cite{iso8800}. Frames are therefore dependent within one part and, if frames are divided at random, also between the selection and the test part. The results are reported for two evaluations and checked with a third run, all with the same seeds and limits. In the \emph{frame-level evaluation}, frames are divided at random into the parts (division by frame, Fig.~\ref{fig:parts}), and each frame is one independent unit. In the \emph{sequence-level evaluation}, whole recorded sequences are divided into the parts (division by recorded sequence), and each sequence is one independent unit. In the sequence-level evaluation, the HB $p$-value and the bounds use, in place of $a_f$, one value per sequence, the mean of $a_f$ over its frames in the condition, and $n$ counts sequences. This is valid if sequences are independent draws, although nuScenes scenes cut from one continuous log share the data-collection vehicle, the day and the weather. Each sequence has equal weight, so this evaluation bounds the mean over sequences of these per-sequence means, which equals $A_c$ if long and short sequences have, on average, the same mean of $a_f$. The third run divides by recorded sequence but counts frames as units, to separate the effect of the division from that of the unit. The 150 nuScenes validation scenes give 15 grid, 68 selection and 67 test scenes, with 9 to 12 rain scenes in each selection and test part. The 45 KITTI validation drives, none of which contributes training frames, give 4 grid, 20 selection and 21 test drives in split~0.

\subsection{Limits, Error Probability and Predictions Fixed in Advance}\label{sec:params}
Before any outcome was computed, the error probability $\delta=0.05$, the limits $\alpha=10^{-2}$, $10^{-3}$ and $10^{-4}$, the conditions and the FAP in the per-frame form of Eq.~(\ref{eq:fap}) were fixed and recorded with a date in version control. These three limits define the performance targets and are called \emph{target limits} (the required limits of the RQs). The \emph{exploratory limits} $\alpha=0.05$ and $0.1$, marked \expl, were fixed at the same time to show at which limits the test reaches the outcome applicable or not applicable. They define no performance target. The sequence-level evaluation and the third run were added afterwards. Predictions of the outcomes at the target limits were recorded together with the outcomes that would count as their failure. The supplementary material lists them with their outcomes.

\section{Results of the Applicability Test on KITTI and nuScenes}\label{sec:results}
Sections~\ref{sec:res-rq1} to~\ref{sec:res-rq3} answer the research questions in turn: the outcomes of the test (RQ1), the changes in the detections that accompany a not applicable outcome (RQ2) and the number of independent units needed (RQ3).
\begin{figure}[t]
\centering
\includegraphics[width=\textwidth]{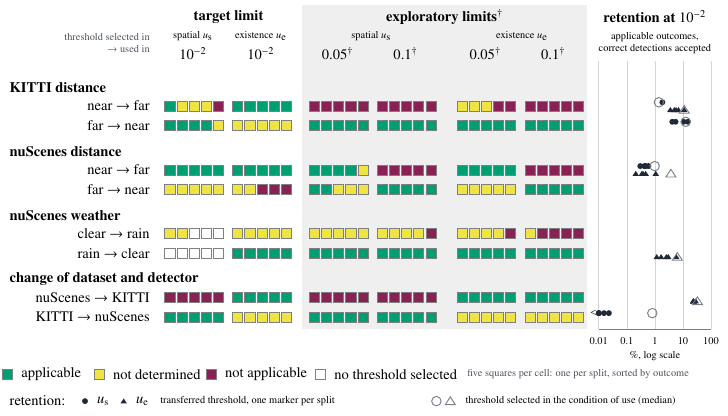}
\Description{A grid with eight rows, two per pair of conditions (KITTI distance, nuScenes distance, nuScenes weather, change of dataset and detector), and six columns: the target limit 0.01 for both indicators on the left, and the exploratory limits 0.05 and 0.1 for both indicators on a grey background. Each cell holds five small squares coloured green (applicable), yellow (not determined), wine red (not applicable) or white (no threshold selected). At 0.01, only the row nuScenes to KITTI for the spatial indicator is wine red on all five splits. At the exploratory limits, the rows near to far, clear to rain and nuScenes to KITTI hold most wine red squares. A panel on the right shows, on a logarithmic axis from 0.01 to 100 percent, the retention of each transferred threshold with an applicable outcome at 0.01, with open markers for the threshold selected in the condition of use.}
\caption{Outcome of every transfer in the frame-level evaluation (left) and retention of the transferred thresholds with an applicable outcome at $10^{-2}$ (right). \expl: exploratory limit, which defines no performance target. `$<$': below 0.01\,\%. No threshold was selected at $10^{-3}$ and $10^{-4}$. Frames of one recorded sequence are not independent, so the stated error probability is not guaranteed.}
\label{fig:outcomes}
\end{figure}

\subsection{Applicability of the Performance Target in the Condition of Use (RQ1)}\label{sec:res-rq1}
Figure~\ref{fig:outcomes} shows all 240 outcomes of the frame-level evaluation, of which 168 were applicable or not applicable. At the target limit $10^{-2}$, the outcome was not applicable on all five splits only after the transfer of the threshold for $\us$ from nuScenes to KITTI. This transfer gave the same outcome at both exploratory limits. After two distance transfers (near to far on KITTI for $\us$, far to near on nuScenes for $\ue$), the outcome at $10^{-2}$ was not applicable on one and on three of five splits. No weather transfer gave this outcome at $10^{-2}$. At 0.1\expl, the outcome was not applicable on at least four of five splits for at least one indicator after transfers into far distance and into rain. In the opposite directions, it was applicable on all five splits.

At $10^{-2}$, transferred thresholds with an applicable outcome had a retention of at most 29\,\% on every split (Fig.~\ref{fig:outcomes}, right). Thresholds selected in the same conditions of use (open markers) reached a median retention of at most 33\,\%. A retention near zero, as for $\us$ from KITTI to nuScenes, means that almost no frame holds an accepted detection, so the limit is met by Eq.~(\ref{eq:trivial}). Every threshold for $\ue$ at $10^{-2}$ lay below 0.25 and accepted only detections found by all four members (Sect.~\ref{sec:detectors}).

Three checks were made. Including the detections excluded by the KITTI rule changed no outcome that held on all five splits. In the third run, every transfer with the outcome not applicable on at least four of five splits in the frame-level evaluation had it again on at least four of five splits. Used in their own condition (Sect.~\ref{sec:rule}), thresholds gave the outcome not applicable in none of the 112 outcomes of the nuScenes distance and weather conditions with a selected threshold, and in 5 of 113 in the third run. Both frequencies lie below the bound $\delta+\delta/2=7.5\,\%$. With the division by frame, frames of one sequence lie in both parts, so the first frequency is weak evidence.

In the sequence-level evaluation, only 5 of the 240 outcomes were applicable or not applicable, all on nuScenes and applicable at 0.1\expl\ (Sect.~\ref{sec:res-rq3}).

\emph{Answer to RQ1.} In the sequence-level evaluation, which the dependence between frames requires, the recorded data show neither that a transferred threshold meets a target limit nor that it does not. In the frame-level evaluation, the outcome at the target limit $10^{-2}$ was not applicable after the transfer of the threshold for $\us$ from nuScenes to KITTI and, on some splits, after a change of distance. After transfers into far distance and into rain, the outcome was not applicable at the exploratory limit 0.1\expl\ for at least one indicator on at least four of five splits. A transferred threshold with an applicable outcome at $10^{-2}$ rejected most correct detections.

\subsection{Changes in the Detections That Accompany a Not Applicable Outcome (RQ2)}\label{sec:res-rq2}
\begin{figure}[t]
\centering
\includegraphics[width=\textwidth]{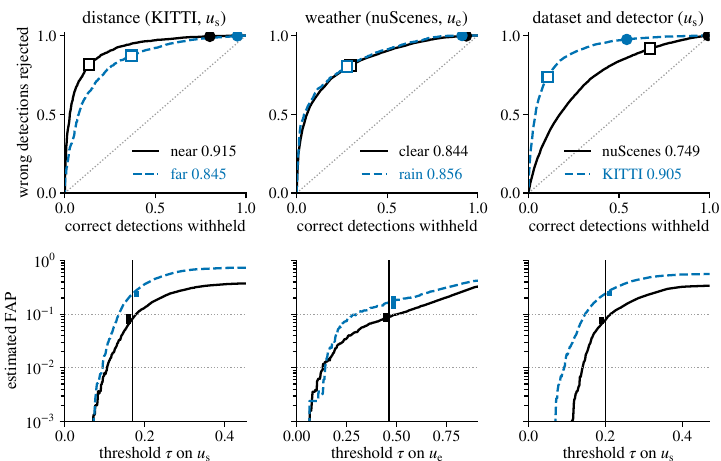}
\Description{Six panels in two rows and three columns: distance (KITTI, spatial indicator), weather (nuScenes, existence indicator), and change of dataset and detector (spatial indicator). Top row: receiver operating characteristic curves of the condition of selection (solid) and the condition of use (dashed), with the transferred threshold marked on both curves. Bottom row: the estimated false acceptance proportion against the threshold for both conditions on a logarithmic axis, with horizontal lines at the limits 0.1 and 0.01, a vertical line at the transferred threshold selected at 0.1, and its interval in each condition.}
\caption{ROC curves (top; legend: AUROC) and estimated FAP against the threshold (bottom) on the test part of split~0, in the condition of selection (solid) and the condition of use (dashed). Squares and vertical lines: transferred threshold selected at 0.1\expl; circles: transferred threshold selected at $10^{-2}$; dotted horizontal lines: limits 0.1\expl\ and $10^{-2}$; bars: interval $[L,U]$.}
\label{fig:separation}
\end{figure}

Except in one transfer, not applicable outcomes occurred only in a condition of use with a larger share of wrong detections than the condition of selection. Table~\ref{tab:causes} lists, for these transfers, the AUROC and the share of wrong detections in both conditions and the further change in the detections. Figure~\ref{fig:separation} shows three of these transfers. AUROC is computed so that a higher indicator value marks a wrong detection, and operating points refer to the test part of split~0.

\begin{table}[t]
\centering
\caption{Changes in the detections from the condition of selection to the condition of use ($\to$) for the transfers into a condition with more wrong detections that gave not applicable outcomes, one indicator each. Shares of wrong detections: all validation frames; other values: test part of split~0, FAP and pooled share at the transferred threshold selected at 0.1\expl. --: no further change reported.}
\label{tab:causes}
\footnotesize
\setlength{\tabcolsep}{3pt}
\begin{tabular}{|L{25mm}|L{19mm}|L{16mm}|L{35mm}|L{15mm}|}
\hline
transfer (indicator) & AUROC & share of wrong detections & further change & shown by AUROC\\
\hline
near $\to$ far, KITTI ($\us$) & 0.915 $\to$ 0.845 & 0.34 $\to$ 0.76 & -- & weaker separation\\
\hline
near $\to$ far, nuScenes ($\us$) & 0.729 $\to$ 0.655 & 0.19 $\to$ 0.51 & -- & weaker separation\\
\hline
clear $\to$ rain, nuScenes ($\ue$) & 0.844 $\to$ 0.856 & 0.33 $\to$ 0.42 & different spread over frames: FAP 0.09 $\to$ 0.17, pooled share 0.12 $\to$ 0.17 & none\\
\hline
nuScenes $\to$ KITTI ($\us$) & 0.749 $\to$ 0.905 & 0.35 $\to$ 0.55 & shifted values: median $\us$ of the correct detections 0.23 $\to$ 0.12 & none\\
\hline
\end{tabular}
\end{table}

With distance, separation weakened and wrong detections became more frequent (Table~\ref{tab:causes}). Among the detections that the threshold selected at near distance accepts, the share of wrong detections grew with distance also below the near/far boundary (supplementary material), so the not applicable outcomes are not specific to this boundary.

In rain, the estimated separation did not weaken. The AUROC of $\ue$ did not decrease, and the transferred threshold selected at 0.1\expl\ marked nearly the same operating point on both curves (Fig.~\ref{fig:separation}). The medians of $\ue$ for correct and for wrong detections moved by at most 0.03, so the values did not shift. The FAP rose more than the pooled share, so the spread of wrong detections over frames added to the effect of more wrong detections. Rain and clear scenes also differ in location and traffic (Sect.~\ref{sec:classes}), so these changes accompany the not applicable outcomes and are not shown to cause them.

Between the datasets, $\us$ took lower values on KITTI (Table~\ref{tab:causes}), so a threshold selected on nuScenes lay high on the KITTI scale. At 0.1\expl, it accepted about 25\,\% of the KITTI wrong detections, against 8\,\% on nuScenes. The outcome was not applicable on KITTI although $\us$ separated correct from wrong detections better there. The threshold is transferred as a raw value, since rescaling per dataset would hide this shift.

The not applicable outcomes after the one transfer into a condition with a smaller share of wrong detections (far to near distance on nuScenes for $\ue$, at $10^{-2}$ on three of five splits) were accompanied by shifted values. The threshold selected at far distance lay higher on the scale than the threshold selected at near distance and therefore accepted more of the wrong detections at near distance.

\emph{Answer to RQ2.} With distance, weaker separation and more wrong detections accompanied the not applicable outcomes. In rain, where not applicable outcomes occurred only at the exploratory limits, more wrong detections and a different spread over frames accompanied them at a nearly unchanged operating point. Between the datasets, shifted values and more wrong detections accompanied them, and shifted values also accompanied the one exception. AUROC shows only weaker separation and therefore showed the not applicable outcomes only for transfers into far distance.

\subsection{Independent Units Needed to Show That a Limit Is Met (RQ3)}\label{sec:res-rq3}
\begin{figure}[t]
\centering
\includegraphics[width=0.86\textwidth]{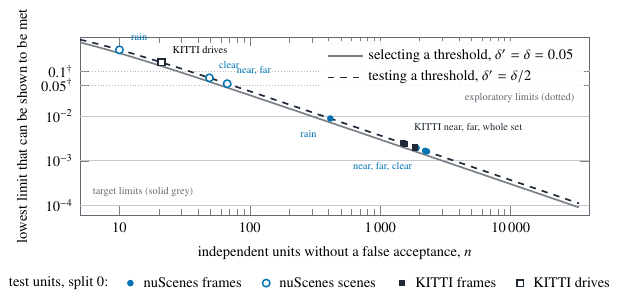}
\Description{A log-log plot of the lowest limit that n independent units without a false acceptance can show to be met, against n, as two nearly parallel falling lines for selecting and testing a threshold. Horizontal lines mark the limits 0.0001, 0.001 and 0.01 (solid grey) and 0.05 and 0.1 (dotted). Filled blue dots for the nuScenes conditions counted in frames lie below the 0.01 line. Open blue circles for the same conditions counted in recorded sequences lie above the 0.01 line, the rain circle above 0.1. Black squares for KITTI frames lie below the 0.01 line, and an open black square for the 21 KITTI test drives lies above the 0.1 line.}
\caption{Lowest limit that $n$ independent units without a false acceptance can show to be met (Eq.~(\ref{eq:samplesize})), with the test units of each condition in split~0. Marker heights are computed from $n$, not measured. Whole set: the KITTI frames of the change of dataset and detector.}
\label{fig:units}
\end{figure}

Figure~\ref{fig:units} places the test units of each condition on the curve of Eq.~(\ref{eq:samplesize}). In the frame-level evaluation, every condition meets the minimum number at $10^{-2}$. At $10^{-3}$, selecting a threshold requires 2{,}995 units without a false acceptance, more than the selection part of any condition holds, so no threshold could be selected at $10^{-3}$ or $10^{-4}$. At $10^{-2}$, no threshold for $\us$ could be selected in rain on any of the five splits, nor in clear weather on three of them, because the selection frames held false acceptances, which raise the number needed above that of Eq.~(\ref{eq:samplesize}).

In the sequence-level evaluation, the 150 nuScenes validation scenes are fewer than the 299 that the selection alone requires at $10^{-2}$. A selection part of 20 to 22 KITTI drives is too small for any limit used here, so no threshold was selected on KITTI. On nuScenes, the scenes of the distance conditions and of clear weather reach the minimum number only at 0.1\expl, and the rain scenes do not reach it.

\emph{Answer to RQ3.} Showing that a limit $\alpha$ is met requires at least the number of independent units of Eq.~(\ref{eq:samplesize}) per condition and part, and more if any unit holds a false acceptance. At $10^{-2}$, these are 299 units for selecting and 368 for testing a threshold. In the frame-level evaluation, KITTI and nuScenes meet this minimum at $10^{-2}$ in every condition and at $10^{-3}$ in none. In the sequence-level evaluation, they can show no target limit to be met, and the rain scenes and the KITTI drives no limit used here.

\section{Discussion}\label{sec:discussion}
This section states what the results mean for a SOTIF analysis and where they may not hold.

\subsection{Consequences for the SOTIF Analysis}\label{sec:implications}

\paragraph{Use of the outcomes in a safety argument.} An outcome is an analysis result for the detector as a system element, not a classification of scenarios, because tracking, fusion and motion planning lie between the detector and a hazardous behaviour. Table~\ref{tab:decision} lists what each outcome means for the performance target and which measure follows. In the condition of selection, the LTT guarantee is the evidence, and a not determined outcome there does not withdraw it. The vehicle manufacturer supplies how often each condition occurs in operation and the limit $\alpha$ from the risk allocated to the detector.

\begin{table}[t]
\centering
\caption{Meaning of each outcome for the performance target in $c'$ and the measure that follows.}
\label{tab:decision}
\footnotesize
\begin{tabular}{|L{20mm}|L{44mm}|L{45mm}|}
\hline
outcome & meaning for the performance target & measure that follows\\
\hline
applicable & evidence, at error probability at most $\delta/2$ if the units are independent, that the target is met in $c'$ with the transferred threshold & use the threshold in $c'$ if the share of annotated cars that reach the motion planner meets a second target on missed objects\\
\hline
not applicable & evidence, at error probability at most $\delta/2$ if the units are independent, that the target is not met in $c'$ with the transferred threshold & detector supplier: select a threshold in $c'$ with a run-time signal for $c'$, or use a common threshold for several conditions; vehicle manufacturer: restrict the ODD or the functionality, or add a measure in tracking, fusion or motion planning\\
\hline
not determined & no statement, because the data of $c'$ do not suffice or the FAP lies close to $\alpha$ & record further independent units (Eq.~(\ref{eq:samplesize}) gives the minimum) and treat $c'$ as not verified\\
\hline
no threshold selected & the lowest candidate threshold fails in $c$ with the available data & record further independent units in $c$ or request a reallocation of $\alpha$ at vehicle level\\
\hline
\end{tabular}
\end{table}

\paragraph{Independent unit and amount of recorded data.} Most transfer outcomes were applicable or not applicable in the frame-level evaluation, almost none in the sequence-level evaluation (Sect.~\ref{sec:res-rq1}). What the data can show therefore depends on the independent unit. It is the recorded sequence unless the independence of frames is shown, because with dependent frames the stated error probability is not guaranteed. Equation~(\ref{eq:samplesize}) gives the minimum number of independent recorded sequences per condition before recording. At $10^{-2}$, this minimum exceeds the validation sets of KITTI and nuScenes, and it grows linearly with $1/\alpha$ and only logarithmically with $1/\delta'$. Recording a rare condition more often than it occurs in operation alters the FAP over all frames (Eq.~(\ref{eq:mix})), but it supplies the units that the test per condition requires. ISO/PAS~8800 notes that such a number of samples is needed to estimate the relative frequency of an error~\cite{iso8800}.

\paragraph{Threshold per condition or common threshold.} A threshold per condition requires the condition to be known at run time. The box gives the distance~\cite{lee2024adaptive}, whereas weather requires a rain signal, whose errors have to be included in the allocation of $\alpha$. A common threshold for several conditions, which has to pass the selection in each of them at error probability $\delta$ divided by their number, needs no run-time signal. For near and far distance, it always gave an applicable or not determined outcome, with a lower retention at near distance than the threshold selected there (supplementary material).

\paragraph{A second performance target on missed objects.} The performance target limits only the wrong detections that are accepted. A missed object can be a more hazardous error than an accepted wrong detection, so a lower threshold exchanges one error for another. A safety argument that uses the threshold therefore needs a second target on missed objects, a lower limit on the share of annotated cars that reach the motion planner. This target can be derived from the same acceptance criterion and tested under the same conditions with the same interval, with Eq.~(\ref{eq:samplesize}) applied to its own limit.

\subsection{Threats to Validity}\label{sec:threats}
\emph{Construct validity.} A wrong detection is defined by the annotations and the overlap criterion, so boxes on annotated cars that miss the overlap criterion and boxes on unannotated objects count as wrong. On KITTI, of the wrong detections that the threshold for $\us$ selected at near distance in Fig.~\ref{fig:teaser}a (limit 0.1\expl, split~0) accepts in all validation frames, 1{,}536 lie inside the camera image, overlap no annotation of another class and overlap no annotated car at an IoU above 0.1. The share of unannotated cars among them was not measured. The near/far boundary is a median of the detector output, so it differs between detectors and splits.

\emph{Internal validity.} Correctness depends on the matching criterion. This paper uses an IoU of 0.5 for both datasets, whereas the KITTI benchmark uses 0.7 for cars and nuScenes matches by the distance between box centres. Matching by this distance would count some far-distance boxes that miss the overlap criterion as correct (not measured). Every detection is anchored on a box of the first member, so a car that this member misses forms no detection even if the other three members detect it. The heading of a merged box is the arithmetic mean of the member headings. This mean can misplace the box when the member headings lie on both sides of $\pm180^{\circ}$ or differ by about $180^{\circ}$, so that the box counts as wrong. The KITTI rule reproduces only part of the ignore logic of the KITTI benchmark, and nuScenes has no equivalent annotation.

\emph{Statistical validity.} Because frames of one recorded sequence are not independent, the frame-level evaluation overstates the evidence. The sequence-level evaluation discards the variation within sequences, and the HB bound does not use the variance of sequence means. The number of effectively independent units is at most the number of frames and can lie below the number of sequences, because nuScenes scenes of one log are not independent either. The AUROC values are point estimates on split~0 without an interval, and in rain they rest on frames of 24 recorded scenes. If frames were independent, the expected number of erroneous applicable or not applicable outcomes among the 240 transfer outcomes of Fig.~\ref{fig:outcomes} would be at most 6 ($240\cdot\delta/2$). Near and far share most KITTI test frames, so the test data of the two conditions are not independent.

\emph{External validity.} One detector architecture, one way of estimating uncertainty (a deep ensemble) and one object class were examined. After the vote, the detections match most annotated cars on KITTI but fewer than half on nuScenes (Table~\ref{tab:setup}), on which the members were trained for about half of the configured epochs. The measured FAPs and retentions are therefore specific to these detectors. Evidence for a product requires the test on data from its ODD, which KITTI and nuScenes do not represent.

\section{Conclusion and Future Work}\label{sec:conclusion}
A detector that meets a performance target by rejecting uncertain detections carries a statistical guarantee only for data like those on which its threshold was selected. A SOTIF analysis, however, has to know whether the target is also met under potential triggering conditions. This paper therefore treats the transfer of a threshold to another condition as a question to be tested. The applicability test answers it at a stated error probability as applicable, not applicable or not determined, and the zero-failure bound gives the minimum number of independent units needed to show that the target is met.

Applied to ensembles of PointPillars networks on KITTI and nuScenes, the test leads to three findings. First, a threshold that keeps the false acceptance proportion within a limit under the condition for which it was selected can exceed this limit under another condition: at the target limit after changes of distance and of dataset, and after a change of weather only at limits above the target limits. AUROC showed this only for transfers into far distance. Second, where a transferred threshold still met the target limit, it did so by rejecting most correct detections, so an applicable outcome is informative only together with its retention, the share of correct detections that the threshold accepts. Third, these findings rest on frames as independent units. Counted in recorded sequences, as the dependence between frames requires, neither validation set is large enough to show a target limit to be met, and no transfer gave a not applicable outcome.

For a SOTIF analysis, a performance target met with an uncertainty threshold therefore counts as evidence under a potential triggering condition only after a test under that condition, reported with the retention and the number of independent units. A not applicable outcome identifies, at the stated error probability and with independent units, a condition in which the detector with this threshold does not meet the target, and Table~\ref{tab:decision} lists the measures that follow. Because the minimum number of independent units can be computed before recording, it can enter the recording plan of each condition.

Future work will extend the test to further potential triggering conditions, among them occlusion and structures that resemble cars, and to further detector architectures and object classes. It will estimate the dependence between frames of one recorded sequence to state the number of effectively independent units, and it will examine thresholds that adapt to the condition at run time. A further direction is to examine whether a monitor of the indicator values in operation, which needs no annotations, detects changes in the detections of the kind found between the datasets and in rain.

\section*{Statements and Declarations}
\paragraph{Funding and competing interests.} This research was funded by the Institute for Driver Assistance and Connected Mobility (IFM), Kempten University of Applied Sciences. The authors have no competing interests relevant to the content of this article.

\paragraph{Data and code availability.}\sloppy\ Analysis scripts, configurations, per-run results and the predictions fixed in advance are available at \url{https://github.com/milinpatel07/lidar-performance-target-applicability}. The KITTI and nuScenes datasets are available from their providers under their own licences.

\paragraph{Use of generative AI.} Generative AI tools were used to assist with the literature screening, the analysis and figure scripts, and the writing of the text. The authors reviewed and verified all content and take full responsibility for this article.

\end{document}